\documentclass[10pt,twocolumn,letterpaper]{article}

\usepackage{cvpr}
\usepackage{times}
\usepackage{epsfig}
\usepackage{graphicx}
\usepackage{amsmath}
\usepackage{amssymb}

\usepackage{bbding}
\usepackage{multirow}
\usepackage{subfigure}
\usepackage{diagbox}
\usepackage{microtype}
\usepackage[american]{babel}

\usepackage[breaklinks=true,bookmarks=false]{hyperref}

\cvprfinalcopy 

\def\cvprPaperID{7555} 

\ifcvprfinal\fi
\begin{document}

\title{SEED: Semantics Enhanced Encoder-Decoder Framework for Scene Text Recognition}

\author{Zhi Qiao$^{1,2}$ \quad Yu Zhou$^1$\thanks{The corresponding author} \quad Dongbao Yang$^1$ \quad Yucan Zhou$^1$ \quad Weiping Wang$^1$\\
$^1$Institute of Information Engineering, Chinese Academy of Sciences, Beijing, China\\
$^2$School of Cyber Security, University of Chinese Academy of Sciences, Beijing, China\\
{\tt\small  \{qiaozhi, zhouyu, yangdongbao, zhouyucan, wangweiping\}@iie.ac.cn}
}

\maketitle
\thispagestyle{empty}

\renewcommand{\thefootnote}{\fnsymbol{footnote}}
\begin{abstract}
   Scene text recognition is a hot research topic in computer vision. Recently, many recognition methods based on the encoder-decoder framework have been proposed, and they can handle scene texts of perspective distortion and curve shape. Nevertheless, they still face lots of challenges like image blur, uneven illumination, and incomplete characters. We argue that most encoder-decoder methods are based on local visual features without explicit global semantic information. In this work, we propose a semantics enhanced encoder-decoder framework to robustly recognize low-quality scene texts. The semantic information is used both in the encoder module for supervision and in the decoder module for initializing. In particular, the state-of-the-art ASTER method is integrated into the proposed framework as an exemplar. Extensive experiments demonstrate that the proposed framework is more robust for low-quality text images, and achieves state-of-the-art results on several benchmark datasets. The source code will be available.\footnote[2]{https://github.com/Pay20Y/SEED}

\end{abstract}

\section{Introduction}

Scene text detection and recognition have attracted great attention in recent years owing to its various applications such as autonomous driving, road sign recognition, helping visual impaired and so on. Inspired by object detection~\cite{liu2016ssd,ren2015faster,lin2017focal,zhang2019freeanchor}, scene text detection~\cite{liao2017textboxes,tian2016detecting,zhou2017east,qin2019curved,chen2019constrained} achieved convincing performance. Despite the maturity of conventional text recognition in documents, scene text recognition is still a challenging task.

\begin{figure}[t]
\begin{center}
\includegraphics[width=0.8\linewidth]{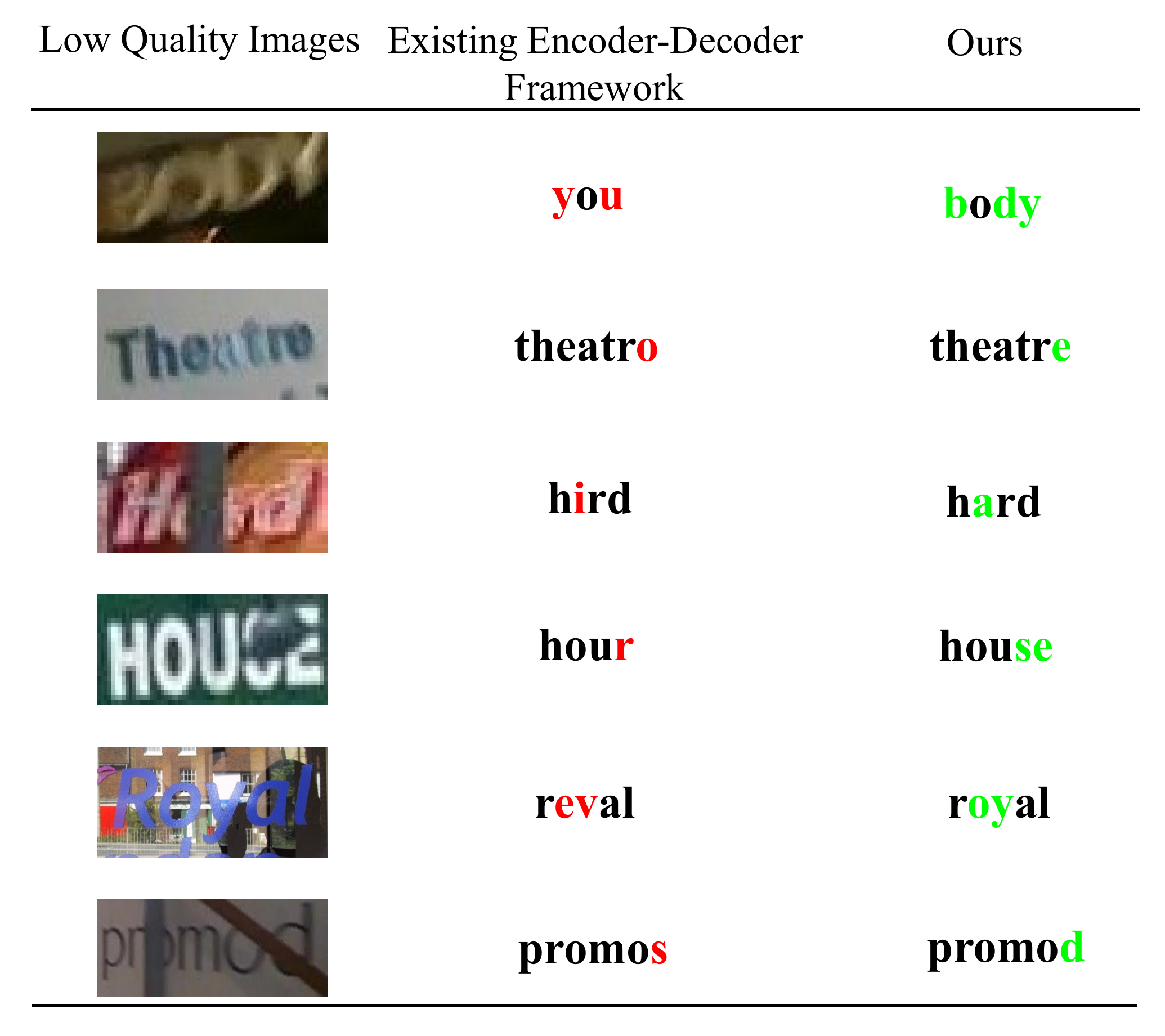}
\end{center}
   \caption{The comparison of our SEED with the existing encoder-decoder framework such as~\cite{shi2018aster}. The first column shows the examples of some challenging scene text including image blur, occlusion, and background interference. The second column is the results of the existing encoder-decoder framework and the third column gives the predictions of our approach. It shows that our proposed method is more robust to the low-quality images. }
   \label{fig_intro_img}
\end{figure}

With the development of deep learning, recent works~\cite{Jaderberg2016Reading,Pan2016Reading,Shi2016An,su2017accurate,lee2016recursive,shi2016robust,shi2018aster,yang2017learning,cheng2017focusing,cheng2018aon,bai2018edit,li2019show,liao2019scene,zhan2019esir,xie2019aggregation,luo2019moran,yang2019symmetry} on scene text recognition have shown promising results. However, existing methods are still facing various problems when dealing with image blur, background interference, occlusion and incomplete characters as shown in Fig.~\ref{fig_intro_img}. 

Recently, inspired by neural machine translation of the natural language processing field, the encoder-decoder framework with attention mechanism has been widely used in scene text recognition. For regular text recognition~\cite{lee2016recursive,cheng2017focusing,ghosh2017visual}, the encoder is based on CNN with RNN and another RNN with attention mechanism is used as the decoder to predict character at each time step. For irregular text recognition, the rectification based methods~\cite{shi2016robust,shi2018aster,liu2018char,zhan2019esir,luo2019moran,yang2019symmetry}, the multi-direction encoding method~\cite{cheng2018aon} and the 2D-attention based methods~\cite{yang2017learning,li2019show} are proposed. Rectification based methods first rectify the irregular images, then the following pipeline is as those of regular recognition. The multi-direction encoding method uses CNN with two LSTMs to encode four different directions. The 2D-attention based methods use 2D-attention mechanism to deal with irregular text which handles feature map from two dimensions directly. 

The existing methods define the text recognition task as a sequence character classification task locally, but ignore the global information of the whole word. As a result, they may struggle to handle low-quality images such as image blur, occlusion and incomplete characters. However, people can deal with these low-quality cases well by considering the global information of the text. 

To address this problem, we propose the \textbf{S}emantics \textbf{E}nhanced \textbf{E}ncoder-\textbf{D}ecoder framework (SEED), in which an additional semantic information is predicted acting as the global information. The semantic information is then used to initialize the decoder as illustrated in Fig.~\ref{fig_encoder_decoder} (c). The semantic information has two main advantages, 1) it can be supervised by a word embedding in natural language processing field, 2) it can reduce the gap between the encoder focusing on the visual feature and the decoder focusing on the language information, since the text recognition can be regarded as a cross-modality task. Specifically, we get the word embedding from a pre-trained language model and compute a loss between the semantic information and the word embedding during training. By this way, the semantic information contains richer semantics, then the predicted semantic information is used to guide the decoding process. As a result, the decoding process can be limited in a semantic space, and the performance of recognition will be better. Some examples are shown in Fig.~\ref{fig_intro_img}. As an example, in the fourth sub-image of Fig.~\ref{fig_intro_img}, the last two characters ``se'' are recognized as ``R'' because of the occlusion, but it can be corrected in our framework with the global semantic information. In other words, the semantic information works as an ``intuition'', which is like a glimpse before people read a word carefully.

Predicting semantic information from images directly has already been studied before.~\cite{gordo2015lewis} predicts semantic concepts directly from a word image with a CNN and a weighted ranking loss.~\cite{wilkinson2016semantic} tries to embed image features into a word embedding space for text spotting.~\cite{krishnan2018word} proposes to learn embedding of the word images and the text labels in an end-to-end way. These works validate that semantic information is helpful to the text related tasks. 

The main contributions are as follows:

1. We propose SEED for scene text recognition, which predicts additional global semantic information to guide the decoding process, and the predicted semantic information is supervised by the word embedding from a pre-trained language model.

2. We integrate the state-of-the-art ASTER method~\cite{shi2018aster} to our framework as an exemplar.

3. Extensive experiments on several public scene text benchmarks demonstrate the proposed framework can obtain state-of-the-art performance, especially on the low-quality datasets ICDAR2015 and SVT-Perspective, and it is particularly more robust for incomplete characters.

The rest of this paper is organized as follows: Sec.~\ref{section_2} reviews the related works, Sec.~\ref{section_3} describes the proposed framework and the exemplar, Sec.~\ref{section_4} conducts profuse experiments and Sec.~\ref{section_5} concludes the work.

\section{Related Work}
\label{section_2}

\subsection{Scene Text Recognition}
Existing scene text recognition methods can be divided into two categories, namely traditional methods and deep learning based methods.

Traditional methods usually adopt a bottom-up approach which detects and classifies characters first and then groups them to a word or text line with heuristic rules, language models or lexicons. They design various hand-craft features then use these features to train a classifier such as SVM. For example,~\cite{neumann2012real} uses a set of computationally expensive features like aspect ratio, hole area ratio, etc.~\cite{wang2010word,wang2011end} use sliding windows with HOG descriptors, and~\cite{yao2014strokelets,bai2016strokelets} use Hough voting with random forest classifier. Most traditional methods suffer from designing various hand-crafted features, and these features are limited for high-level representation.

With the development of deep learning, most methods use CNN to perform a top-down approach which recognizes word or text line directly.~\cite{Jaderberg2016Reading} treats a word as a class, then converts the recognition problem into the image classification problem. Recently, most works treat the recognition problem as the sequence prediction problem. Existing methods can be almost divided into two techniques namely Connectionist Temporal Classiﬁcation (CTC) and attention mechanism. For CTC-based decoding,~\cite{Pan2016Reading,Shi2016An,su2017accurate} propose to use CNN and RNN to encode the sequence features and use CTC for character alignment. For attention-based decoding,~\cite{lee2016recursive} proposes recursive CNN to capture longer contextual dependencies and uses an attention-based decoder for sequence generation.~\cite{cheng2017focusing} introduces the problem of attention drift, and proposes focusing attention for better performance. 

However, these works all assume that the text is horizontal, and can not handle the text of irregular shapes such as perspective distortion and curvature. To solve the problem of irregular text recognition,~\cite{shi2016robust,shi2018aster} propose to rectify the text first based on Spatial Transformer Network~\cite{jaderberg2015spatial} and then treat it as horizontal text. Furthermore,~\cite{zhan2019esir} gets better performance with iterative rectiﬁcation and~\cite{yang2019symmetry} rectifies with some geometric constraints.~\cite{luo2019moran} rectifies text by predicting pixels offset. Instead of rectifying the whole text,~\cite{liu2018char} takes an approach of detecting and rectifying individual characters. In spite of rectification,~\cite{cheng2018aon} encodes the images in four directions and proposes a filter gate to fuse the features.~\cite{yang2017learning} introduces an auxiliary dense character detection task and an alignment loss into the 2D attention based network.~\cite{li2019show} proposes a tailored 2D attention based framework for irregular text recognition. Without encoder-decoder framework,~\cite{liao2019scene} converts irregular text recognition into character segmentation with fully convolutional network~\cite{long2015fully}.~\cite{xie2019aggregation} proposes a new loss function for more effective decoding. 

\subsection{Semantics in Scene Text}
Many works try to bring semantics into the tasks of text recognition or text retrieval.~\cite{gordo2015lewis} predicts semantic concepts directly from a word image with CNN.~\cite{patel2016dynamic} proposes to generate contextualized lexicons for scene images with only visual information, and word-spotting task benefits a lot from the lexicons.~\cite{wilkinson2016semantic,krishnan2018word} learn to map the word images to a word embedding space and apply it into word spotting system.~\cite{kang2017detection} tries to detect and recognize text in online images with the help of context information such as tags, comments, and titles.~\cite{sabir2018visual} introduces to use the language model and the semantic correlation between scene and text to re-rank the recognition results.~\cite{prasad2018using} proposes to boost the performance of text spotting with the object information.~\cite{ghosh2018don} uses the text embedded in advertisement images to enhance the image classiﬁcation.~\cite{zheng2019deep} proposes to use a pre-trained language model to correct the inaccurate recognition results with the text context in the image.

As discussed before, state-of-the-art recognition methods do not utilize the semantics of the text well. The related semantics works do not integrate the semantics into the recognition pipeline explicitly and effectively.


\section{Method}
\label{section_3}

\begin{figure}[t]
\begin{center}
   \subfigure[Plain Encoder-Decoder Framework]{
      \includegraphics[width=1.0\linewidth]{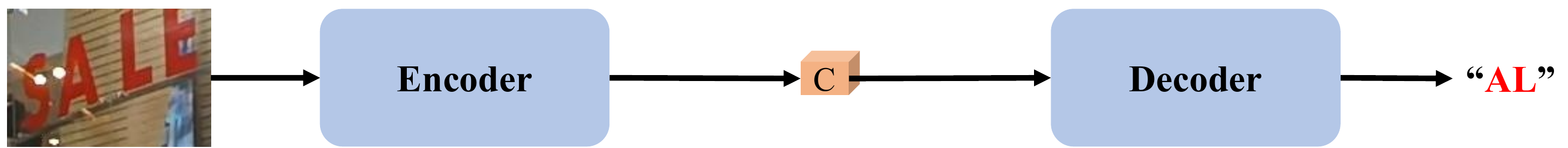}
   }
   \subfigure[Attention-Based Encoder-Decoder Framework]{
      \includegraphics[width=1.0\linewidth]{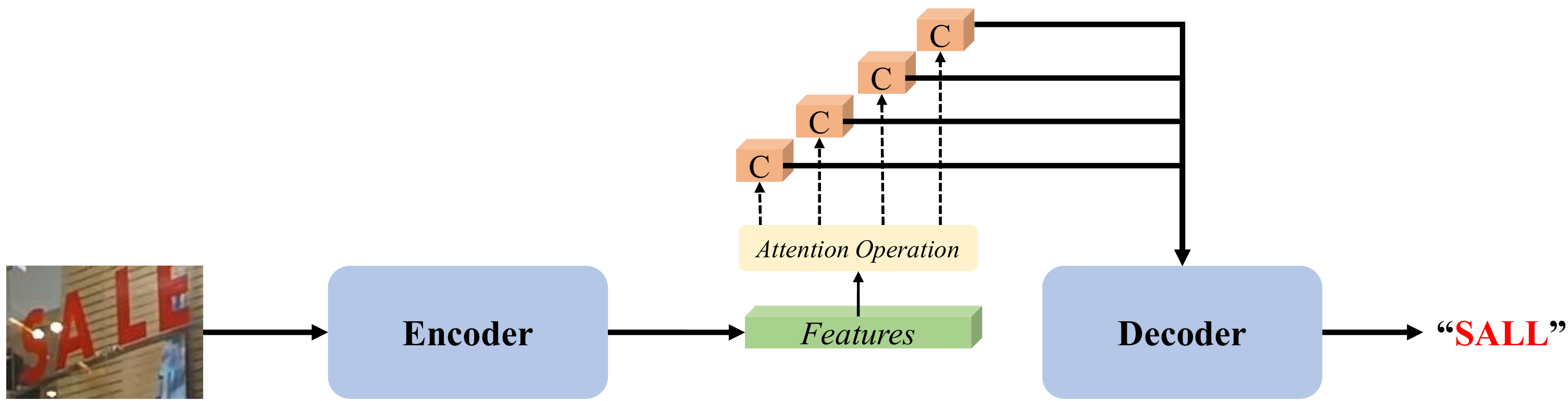}
   }
   \subfigure[Our Encoder-Decoder Framework]{
      \includegraphics[width=1.0\linewidth]{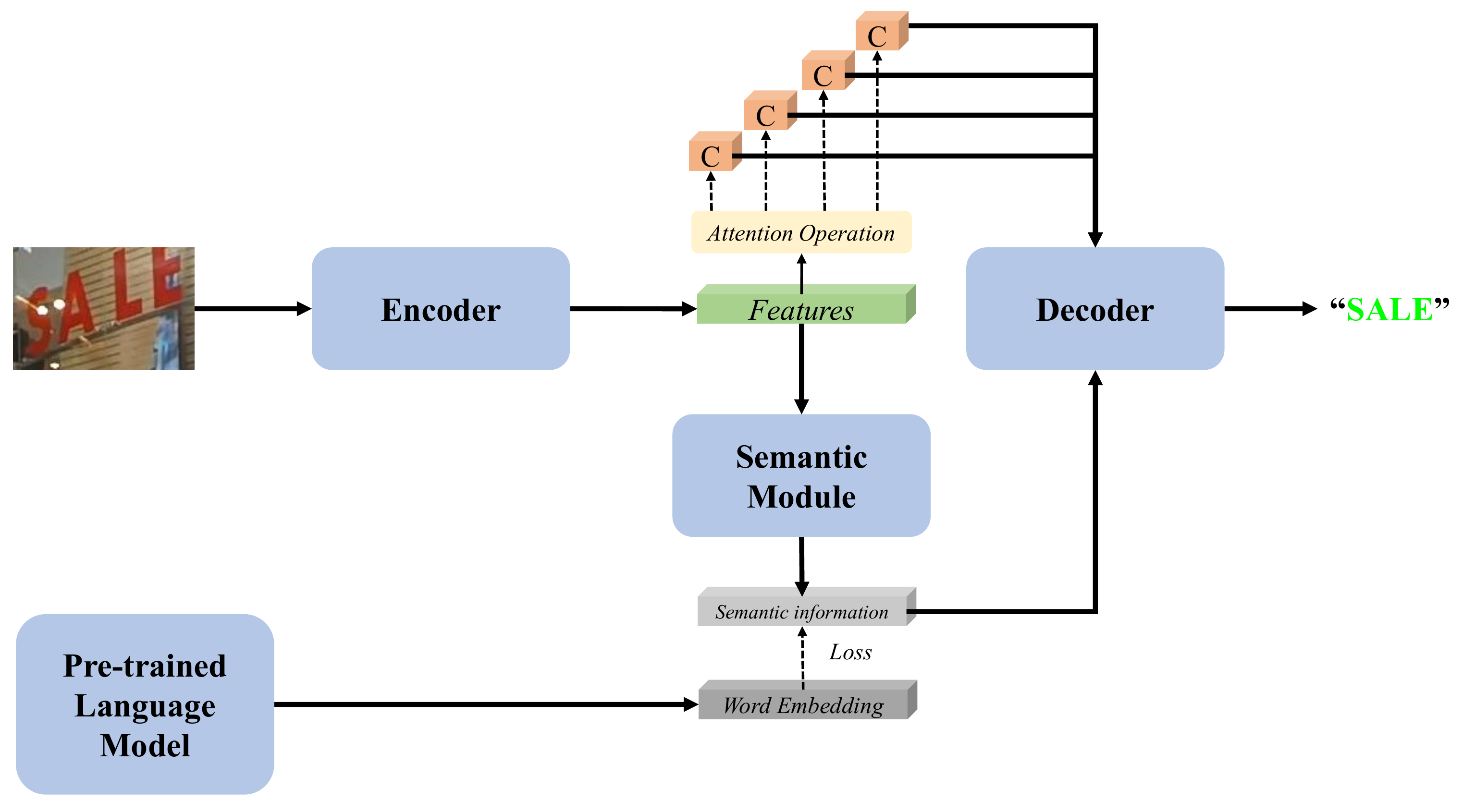}
   }
\end{center}
   \caption{Comparison of three kinds of framework. ``C'' represents context information. The plain encoder-decoder framework gets incorrect results due to limited context representation. The attention-based encoder-decoder framework works better but still can not handle incomplete characters without global information. Our proposed encoder-decoder framework predicts the correct result with the help of global semantic information.}
\label{fig_encoder_decoder}
\end{figure}

In this section we describe the proposed method in detail. The general framework is shown in Fig.~\ref{fig_encoder_decoder} (c), which consists of 4 major components: 1) The \textit{encoder} including CNN backbone and RNN for extracting visual features; 2) The \textit{semantic module} for predicting semantic information from the visual features; 3) The \textit{pre-trained language model} for supervising the semantic information predicted by \textit{semantic module}; 4) The \textit{decoder} including RNN with attention mechanism for generating the recognition results. First we review the encoder-decoder framework in Sec.~\ref{section3_1}, and introduce the pre-trained language model detailedly in Sec.~\ref{section3_2}. In Sec.~\ref{section3_3}, we describe our proposed method. Specifically, we present the general framework in Sec.~\ref{section3_3_1}. After that, we show the details of the proposed method which integrate state-of-the-art method ASTER~\cite{shi2018aster} into proposed framework in Sec.~\ref{section3_3_2}. Finally, the loss function and the training strategies are presented in Sec.~\ref{section3_4}.

\subsection{Encoder-Decoder Framework}
\label{section3_1}

The Encoder-decoder framework is widely used in neural machine translation, speech recognition, text recognition and so on.~\cite{sutskever2014sequence} first introduces the structure of the framework and applies it into neural machine translation. For simplicity, we call this framework plain encoder-decoder framework. As visualized in Fig.~\ref{fig_encoder_decoder} (a), the encoder extracts rich features and generates a context vector $ C $ which contains global information of the inputs, then the decoder converts the context vector to target outputs. Source inputs and target outputs are different due to different tasks, as for text recognition, the inputs are images and target outputs are the texts in the images. The specific composition of encoder and decoder is not fixed, CNN and LSTM are all common choices.

Despite great effectiveness, the plain encoder-decoder framework has an obvious drawback, where the context information has limited ability to represent the whole inputs. Inspired by human visual attention, researchers introduce the attention mechanism into the encoder-decoder framework, which is defined as the attention-based encoder-decoder framework. As shown in Fig.~\ref{fig_encoder_decoder} (b), attention mechanism attempts to build shortcuts between the context and the whole inputs. The decoder can select the appropriate context at each decoding step which is capable of resolving long-range dependency problems, and the alignment between encoder and decoder is trained in a weakly supervised way. 

For scene text recognition, the decoder only depends on the limited local visual features for decoding in both the plain encoder-decoder framework and the attention-based encoder-decoder framework, so it is difficult to deal with some low-quality images without global information. In our proposed framework, the encoder learns explicit global semantic information and uses it as guidance for the decoder. We use FastText~\cite{bojanowski2017enriching} to generate word embedding as the supervision of the semantic information in that it can solve the problem of ``out of vocabulary''.

\subsection{FastText Model}
\label{section3_2}

We choose FastText as our pre-trained language model, which is based on skip-gram. Let $ T = \{w_{i-l}, \ldots, w_{i+l}\} $ be a sentence in a text corpus. $ l $ indicates the length of the sentence and is a hyper-parameter. In skip-gram, a word $ w_i $ is represented by a single embedding vector $ v_i $ and then input to a simple feed-forward neural network, which aims to predict the context represented as $ C_i = \{w_{i-l}, \ldots, w_{i-1},w_{i+1}, \ldots, w_{i+l}\} $. With training the feed-forward network, the embedding vector is simultaneously optimized, and the final embedding vector of a word is close to the words with similar semantics.

FastText additionally embeds subwords and uses them to generate final embedding of the word $ w_i $. Given the hyperparamters $ l_{min} $ and $ l_{max} $ denoting a minimum and a maximum length of the subwords. For example, let $l_{min}=2$, $l_{max}=4$ and the word be ``where'', the set of subwords is \{\textit{wh}, \textit{he}, \textit{er}, \textit{re}, \textit{whe}, \textit{her}, \textit{ere}, \textit{wher}, \textit{here}\}. The word representation is obtained by the combination of the embedding vectors of all subwords and the word itself. Accordingly, FastText model can handle the problem of ``out of vocabulary''. There are some novel words or incomplete words in the benchmark datasets such as ICDAR2015 and SVT-Perspective, so FastText is suitable for our framework. 

\subsection{SEED}
\label{section3_3}

\subsubsection{General Framework}
\label{section3_3_1}

Many scene text recognition methods are based on the encoder-decoder framework with attention. The decoder focuses on specific regions of visual features and outputs corresponding characters step by step. The framework works well in most scenarios except in low-quality images. In some low-quality images, texts may be blurred or occluded. To address these problems, utilizing global semantic information is an alternative. The proposed framework is shown in Fig.~\ref{fig_encoder_decoder} (c). Different from the attention-based encoder-decoder framework, the proposed semantic module predicts extra semantic information. Further, we use the word embedding from a pre-trained language model as the supervision to improve the performance. After that, the semantic information is fed into the decoder along with the visual features. In this way, our method is robust to low-quality images and can correct recognition mistakes. 

\subsubsection{Architecture of Semantics Enhanced ASTER}
\label{section3_3_2}
\begin{figure}[t]
\begin{center}
\includegraphics[width=1.0\linewidth]{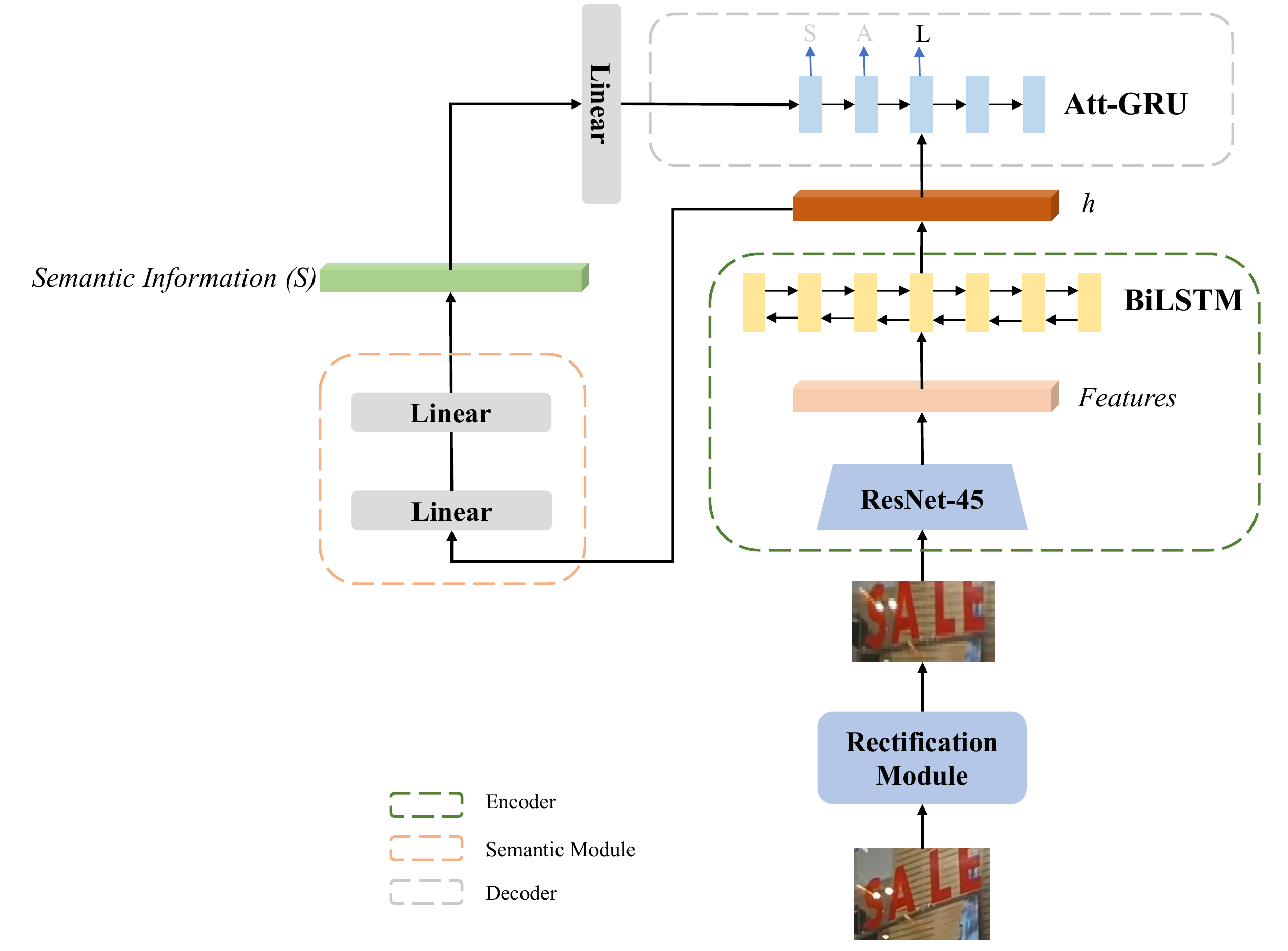}
\end{center}
   \caption{Details of our SE-ASTER. It consists of four main modules, rectification module, encoder, semantic module, and decoder. The semantic module predicts semantic information from the outputs of the encoder which is fed into decoder as the guidance. }
   \label{fig_se_aster}
\end{figure}

We use ASTER~\cite{shi2018aster} as an exemplar for our proposed framework, and we call the proposed method \textbf{S}emantics \textbf{E}nhanced \textbf{ASTER} (SE-ASTER). The SE-ASTER is illustrated in Fig.~\ref{fig_se_aster}. There are four modules: the rectification module is to rectify the irregular text images, the encoder is to extract rich visual features, the semantic module is to predict semantic information from the visual features, and the decoder transcribes the final recognition results. 

First, the image is input to the rectification module to predict control points with a shallow CNN, then Thin-plate Splines~\cite{warps1989thin} is applied to the image. In this way, the distorted text image will be rectified. This module is the same as~\cite{shi2018aster}, so we don't describe it in detail. Thereafter, the rectified image will be input to the encoder, and rich visual features can be generated. Specifically, the encoder consists of a 45-layer ResNet based CNN same as~\cite{shi2018aster} and a 2-layer Bidirectional LSTM~\cite{graves2008novel} (BiLSTM) network with 256 hidden units. The output of the encoder is a feature sequence $ h = (h_1, \ldots, h_L) $ with the shape of $ L \times C $, where $ L $ is the width of the last feature map in CNN, and $ C $ is the depth. 

The feature sequence $ h $ has two functions, one is to predict the semantic information by the semantic module and the other is as the input of the decoder. For predicting semantic information, we first flatten the feature sequence into a one-dimensional feature vector $ I $ with dimension of $ K $, where $ K = L \times C $. The semantic information $ S $ is predicted with two linear functions as following: 
\begin{equation}
S = W_2\sigma(W_1I + b_1) + b_2\,.
\end{equation}
where $ W_1, W_2, b_1, b_2 $ are trainable weights in the linear function, $ \sigma $ is a ReLU activation function. We also evaluate predicting the semantic information with the final hidden state $ h_L $ of BiLSTM in the encoder, and it gets worse performance. It may originate from that predicting semantic information needs larger feature contexts and it is more proper to use the BiLSTM outputs. The semantic information will be supervised by the word embedding provided by the pre-trained FastText model. The loss function used here will be introduced in Sec.~\ref{section3_4}. 

The decoder adopts the Bahdanau-Attention mechanism~\cite{bahdanau2014neural} which consists of a single layer attentional GRU~\cite{cho2014learning} with 512 hidden units and 512 attention units. Different from~\cite{shi2018aster} we use a single direction decoder here. In particular, the semantic information $ S $ is used to initialize the states of GRU after a linear function for transforming the dimension. Instead of using zero-state initializing, the decoding process will be guided with global semantics, so the decoder uses not only local visual information but also global semantic information to generate more accurate results. 

\subsection{Loss Function and Training Strategy}
\label{section3_4}

We add supervision at both the semantic module and the decoder module. SE-ASTER is trained end-to-end. The loss function is as follows:

\begin{equation}
\label{eq_2}
L = L_{rec} + \lambda L_{sem}\,.
\end{equation}
where $ L_{rec} $ is the standard cross-entropy loss of the predicted probabilities with respect to the ground-truth, and $ L_{sem} $ is the cosine embedding loss of the predicted semantic information with respect to the word embedding of the transcription label from the pre-trained FastText model. $ \lambda $ is hyper-parameters to balance the loss, and we set it to 1 here. Note that we just use a simple cosine based loss function here instead of contrastive loss for faster training speed. 

\begin{equation}
L_{sem} = 1 - cos(S, em)\,.
\end{equation}
where $ S $ is the predicted semantic information and $ em $ is the word embedding from pre-trained FastText model. 

There are two training strategies. The first is initializing the state of the decoder with the word embedding from the pre-trained FastText model rather than the predicted semantic information. Another is to use the predicted semantic information directly. We evaluate these two strategies, and their performances are similar. We use the second training strategy which trains the model in a pure end-to-end way. 


\section{Experiments}
\label{section_4}

In this section, we conduct extensive experiments to verify the effectiveness of our proposed method. First, we introduce the datasets used for training and evaluation, and the implementation details are described. Next, we perform ablation studies to analyze the performance of the different strategies. Finally, our method is compared with state-of-the-art methods on several benchmarks. 


\subsection{Datasets}

\textbf{IIIT5K-Words (IIIT5K)}~\cite{mishra2012scene} contains 5000 images, most of which are regular samples. There are 3000 images for testing. Each sample in test set is associated with a 50-word lexicon and a 1k-word lexicon.

\textbf{Street View Text (SVT)}~\cite{wang2011end} consists of 647 cropped word images from 249 street view images. Most of word images are horizontal, but some of them are severely corrupted by noise, blur, and low resolution. A 50-word lexicon is provided for each image.

\textbf{SVT-Perspective (SVTP)}~\cite{quy2013recognizing} contains 645 word images for evaluation. most images suffer in heavy perspective distortions which are difficult for recognition. Each image is associated with a 50-word lexicon.

\textbf{ICDAR2013 (IC13)}~\cite{karatzas2013icdar} consists of 1015 images for testing, most of which are regular text images. Some of them are under uneven illumination.

\textbf{ICDAR2015 (IC15)}~\cite{karatzas2015icdar} was collected without careful capture. Most of images are with various distortions and blurry which are challenging for most existing methods.

\textbf{CUTE80 (CUTE)}~\cite{risnumawan2014robust} consists of 288 word images only for evaluation. Most of them are curved but with high resolution, no lexicon is provided.

\textbf{Synth90K}~\cite{Jaderberg2016Reading} consists of 9 million synthetic images generated from a lexicon of 90K words. It has been widely used in text recognition task. We use it as one of our training datasets. It contains words from the testing set of the IC13 and SVT.

\textbf{SynthText}~\cite{gupta2016synthetic} is another synthetic dataset for text detection task. We crop the words with ground-truth word bounding boxes and use for training our model.

\begin{table*}[h]
\begin{center}
\small
   \begin{tabular}{|c|c|c|c|c|c|c|c|c|}
   \hline
   & \multicolumn{4}{|c|}{\textbf{IC13-sr}} & \multicolumn{4}{|c|}{\textbf{IC15-sr}} \\
   \hline
   \diagbox{\textbf{Methods}}{\textbf{Images}} & \includegraphics[scale=0.2]{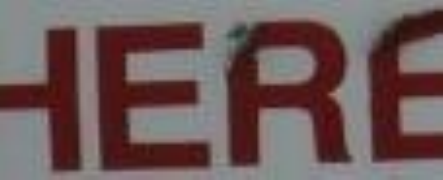} & \includegraphics[scale=0.2]{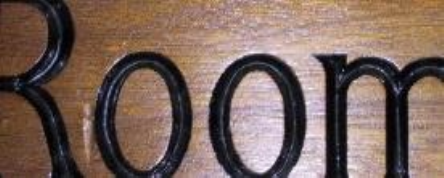} & \includegraphics[scale=0.2]{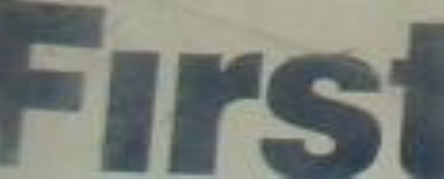} & \includegraphics[scale=0.2]{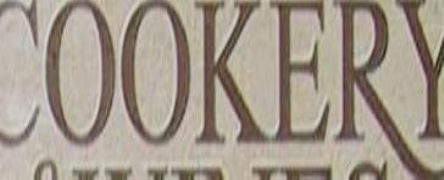} & \includegraphics[scale=0.2]{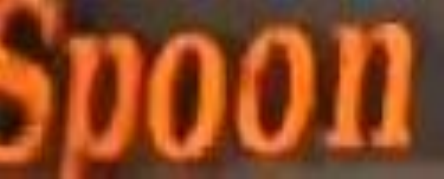} & \includegraphics[scale=0.2]{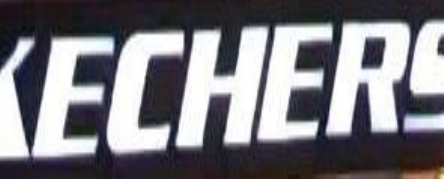} & \includegraphics[scale=0.2]{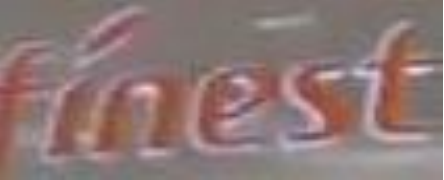} & \includegraphics[scale=0.2]{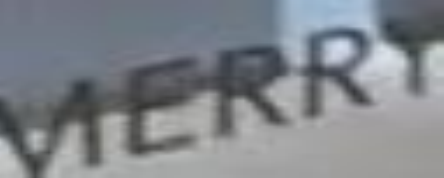} \\
   \hline
   \textbf{ASTER} & {\color{red} ier} & {\color{red} for} & {\color{red} irst} & {\color{red} cooker} & {\color{red} poon} & {\color{red} echars} & {\color{red} mest} & {\color{red} xerr} \\
   \hline
   \textbf{ASTER+WES} & {\color{red} herb} & {\color{green} room} & {\color{red} its} & {\color{red} ooker} & {\color{green} spoon} & {\color{red} rechery} & {\color{red} inest} & {\color{red} verry} \\
   \hline
   \textbf{ASTER+INIT} & {\color{green} here} & {\color{red} look} & {\color{green} first} & {\color{red} looker} & {\color{red} poon} & {\color{red} keekers} & {\color{red} inest} & {\color{red} merri} \\
   \hline
   \textbf{SE-ASTER} & {\color{green} here} & {\color{green} room} & {\color{green} first} & {\color{green} cookery} & {\color{green} spoon} & {\color{green} kechers} & {\color{green} finest} & {\color{green} merry} \\
   \hline
   \end{tabular}
\end{center}
\caption{Visualization of the recognition results on the two shrink datasets. Red: wrong results; Green: correct results. }
\label{tabel_vis_sr}
\end{table*}

\begin{table}[!h]
\begin{center}
\small
   \begin{tabular}{c|cc|c|c|c}
   \hline Methods & WES        & INIT       & IC13          & SVTP          & IC15 \\
   \hline ASTER~\cite{shi2018aster}   &            &            & 91.8          & 78.5          & 76.1 \\
   ASTER-r        &            &            & 90.9          & 79.1          & 78.4 \\
   ASTER          & \Checkmark &            & 90.8          & 79.2          & 77.0 \\
   ASTER          &            & \Checkmark & 91.1          & 78.1          & 76.1  \\
   ASTER          & \Checkmark & \Checkmark & \textbf{92.8} & \textbf{81.4} & \textbf{80.0} \\
   \hline  
   \end{tabular}
\end{center}
\caption{Performance comparison between different strategies. WES represents word embedding supervision. INIT represents initializing the state of the GRU in the decoder. ASTER-r represents the model re-trained by ourselves.}
\label{tabel_ablation}
\end{table}

\subsection{Implementation Details}
\renewcommand{\thefootnote}{\arabic{footnote}}
The proposed SE-ASTER is implemented in PyTorch~\cite{NIPS2019_9015}. The pre-trained FastText model is the officially available model\footnote{https://fasttext.cc/docs/en/crawl-vectors.html} trained on \textit{Common Crawl}\footnote{https://commoncrawl.org/} and \textit{Wikipedia}\footnote{https://www.wikipedia.org/}. In total 97 symbols are recognized, including digits, upper-case and lower-case letters, 32 punctuation marks, end-of-sequence symbol, padding symbol, and unknown symbol.

The size of input images are resized to $ 64 \times 256 $ without keeping ratio, and we adopt the ADADELTA~\cite{zeiler2012adadelta} to minimize the objective function. Without any pre-training and data augmentation, our model is trained on SynthText and Synth90K for 6 epochs with the batch size of 512, the learning rate is set to 1.0 and is decayed to 0.1 and 0.01 at the 4th epoch and the 5th epoch. The model is trained on one NVIDIA M40 graphics card. 

For evaluation, we resize the input images to the same size as for training. We use beam search for GRU decoding, which keeps the $ k $ candidates with the highest accumulative scores, where $ k $ is set to $ 5 $ in all our experiments.

\subsection{Ablation Study}

There are two steps about the semantic module, one is the word embedding supervision and the other is initializing decoder with the predicted semantic information. We evaluate these two steps separately by using the Synth90K and SynthText as training data consistently. The results are shown in Tab.~\ref{tabel_ablation}. The model supervised with word embedding only does not improve the performance compared with the baselines. Using predicted holistic features from the encoder to initialize decoder improves the performance by almost $ 0.2\% $ in ICDAR13, but gets worse performance on SVTP and IC15. It shows that learning global information in an implicit weakly supervised way still struggles with low-quality images. A combination of these two steps gets the best performance. The improvements of $ 1.9\% $, $ 2.3\% $ and $ 1.6\% $ are obtained on IC13, SVTP and IC15 respectively. Compared with ASTER without word embedding supervision, it improves the accuracy by $ 1.7\% $ on IC13, $ 3.3\% $ on SVTP and $3.9\% $ on IC15, which verifies that the supervision with word embedding is quite important. 

\subsection{Performance with Inaccurate Bounding Boxes}

Scene text recognition in real applications is always combined with the detection branch to achieve an end-to-end pipeline. However, the detection branch may not output ideal bounding boxes. If text recognition is robust to inaccurate detection results, the overall end-to-end performance can be more satisfactory. Limited by the receptive field of CNNs, the most frequent inaccurate detection is incomplete characters. We conduct experiments to show our method is robust with this situation. Here we also use SE-ASTER as an exemplar. Note that the SE-ASTER is only trained on Synth90K and SynthText without any data augmentation such as random cropping. We first generate two shrink datasets IC13-sr and IC15-sr based on IC13 and IC15 respectively as follows.

We randomly remove the original word images up to $ 15\% $ in the left, right, top and bottom directions simultaneously. All of the cropped images still have an intersection over union with the original ones larger or equal than $ (1 - 0.15 \times 2)^2 = 0.49 $. According to the evaluation protocol of detection, these cropped images are all positive localizations because the IoU is above the standard threshold of 0.5. Some examples are shown in Tab.~\ref{tabel_vis_sr}.

\newcommand{\tabincell}[2]{\begin{tabular}{@{}#1@{}}#2\end{tabular}}  

\begin{table}[h]
\begin{center}
\small
   \begin{tabular}{l|cc|cc}
   \hline
   Methods & \tabincell{c}{IC13 \\ IC13-sr} & GAP & \tabincell{c}{IC15 \\ IC15-sr} & GAP \\ 
   \cline{2-5}
   \hline
   ASTER      & \tabincell{c}{90.9 \\ 71.4} & -19.5          & \tabincell{c}{78.4 \\ 65.6} & -12.8 \\
   \hline
   ASTER+WES  & \tabincell{c}{90.8 \\ 71.9} & -18.9          & \tabincell{c}{77.0 \\ 62.8} & -14.2 \\
   \hline
   ASTER+INIT & \tabincell{c}{91.1 \\ 74.6} & -16.5          & \tabincell{c}{76.1 \\ 63.1} & -13.0 \\
   \hline
   SE-ASTER    & \tabincell{c}{92.8 \\ 77.4} & \textbf{-15.4} & \tabincell{c}{80.0 \\ 70.0} & \textbf{-10.0} \\
   \hline
   \end{tabular}
\end{center}
\caption{Results on the shrink datasets, GAP indicates the decline between two datasets.}
\label{tabel_sr}
\end{table}

The quantitative results are illustrated in Tab.~\ref{tabel_sr}. The performances of the ASTER baseline drop $ 19.5\% $ and $ 12.8\% $ on the IC13-sr dataset and the IC15-sr dataset respectively, which reveal that the ASTER baseline suffers a lot from the incomplete characters. However, with the supervision of word embedding, the model still struggles with the shrink images. Using the holistic information from encoder as the guidance of the decoder gets better results with $ 16.5\% $ and $ 13.0\% $ decline. SE-ASTER gets the best results, which shows that our model is more robust with incomplete characters. Some visualizing examples are illustrated in Tab.~\ref{tabel_vis_sr}.

\begin{figure}[t]
\begin{center}
   \includegraphics[width=1.0\linewidth]{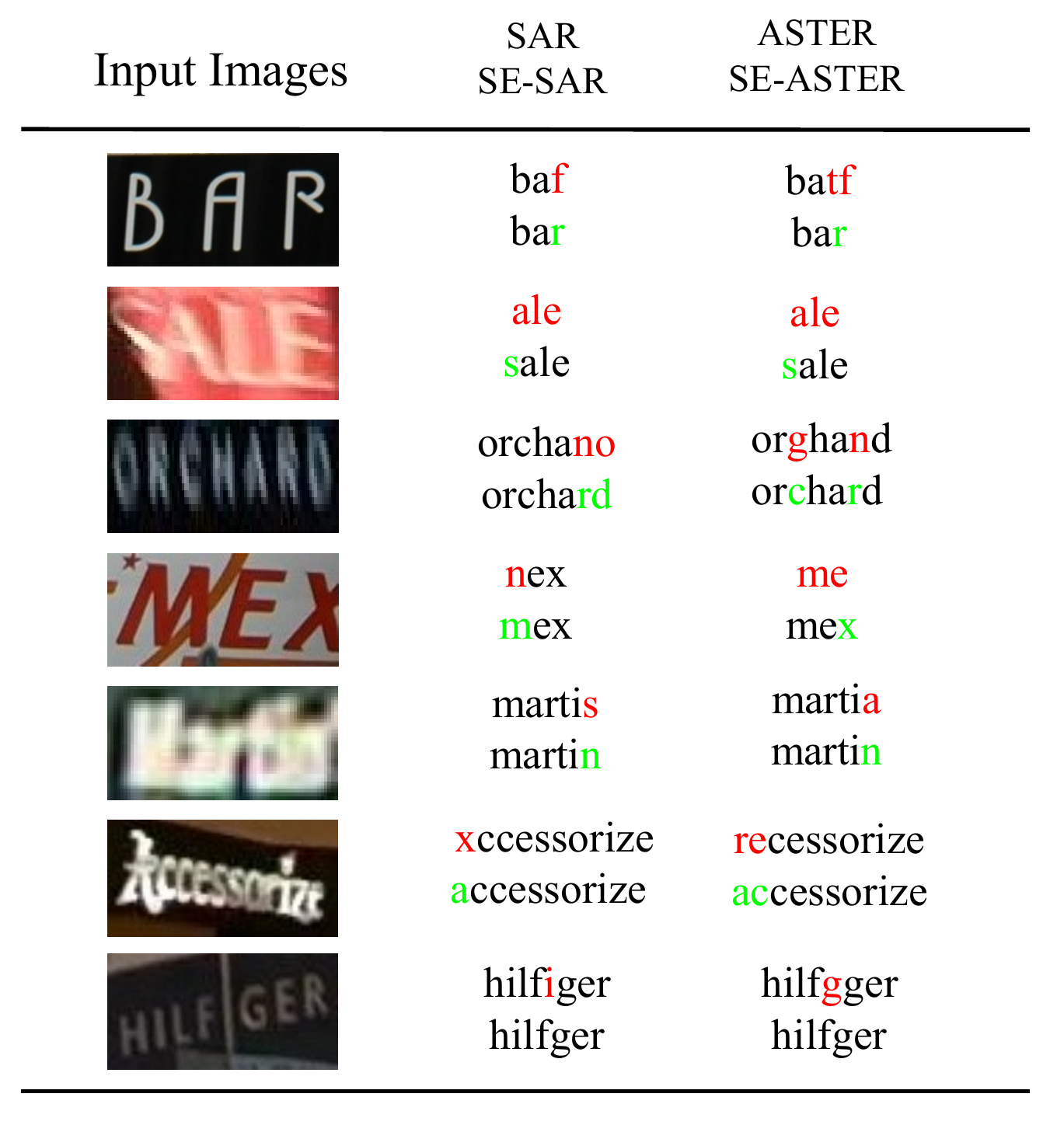}
\end{center}
   \caption{Examples of low-quality images and recognition results in four methods. Red characters are the wrong results, and green ones are the correct. }
   \label{fig_low_quality}
\end{figure}

\subsection{Generalization of Proposed Framework}

To verify the generalization of SEED, we integrate another state-of-the-art recognition method SAR~\cite{li2019show}. SAR is a 2D-attention based recognition method without rectification on input images, and it already adopts an LSTM to generate a holistic feature. However as we mentioned before, the holistic feature may be not effective in a weakly supervised training strategy, so we make some modifications and call our new model \textbf{S}emantics \textbf{E}nhanced \textbf{SAR} (SE-SAR). 

In SE-SAR, we replace the max-pooling along the vertical axis with a shallow CNN. The output of the shallow CNN is a feature map with the height of $ 1 $, then the feature map is fed into a 2-layer LSTM to extract context information. Two linear functions are applied to the output of LSTM to predict the semantic information. Except for the 2D-attention decoder in SAR, we apply another decoder to the output of the LSTM and supervise with the transcription labels. In this way, the output of LSTM contains richer information and helps predict semantic information. Finally, the semantic information is used to initialize the LSTM of the decoder. The model is trained on Synth90K and SynthText for 2 epochs with the batch size of 128. 

\begin{figure}[h]
\centering
   \subfigure[]{
      \includegraphics[scale=0.3]{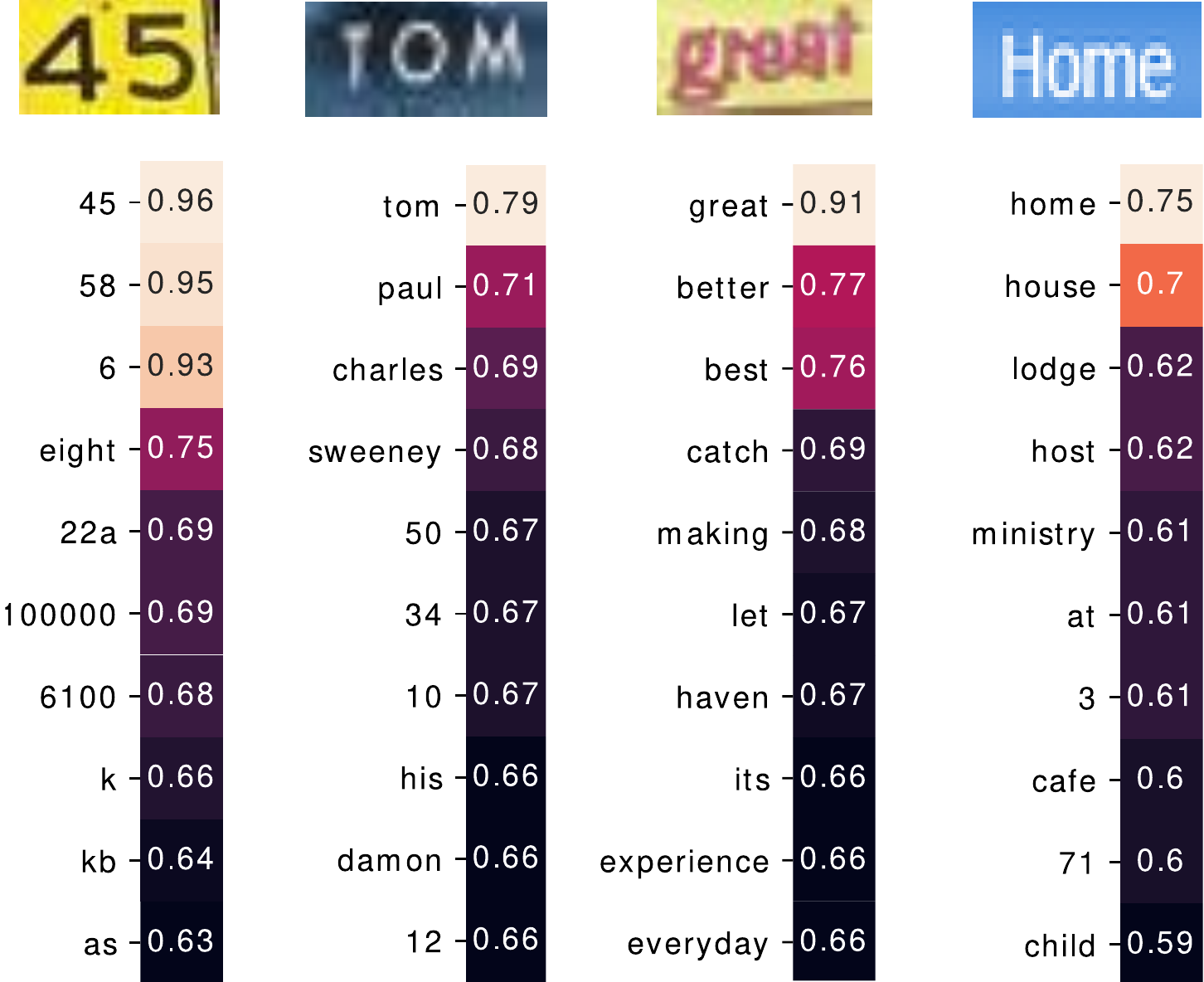}
   }

   \subfigure[]{
      \includegraphics[scale=0.3]{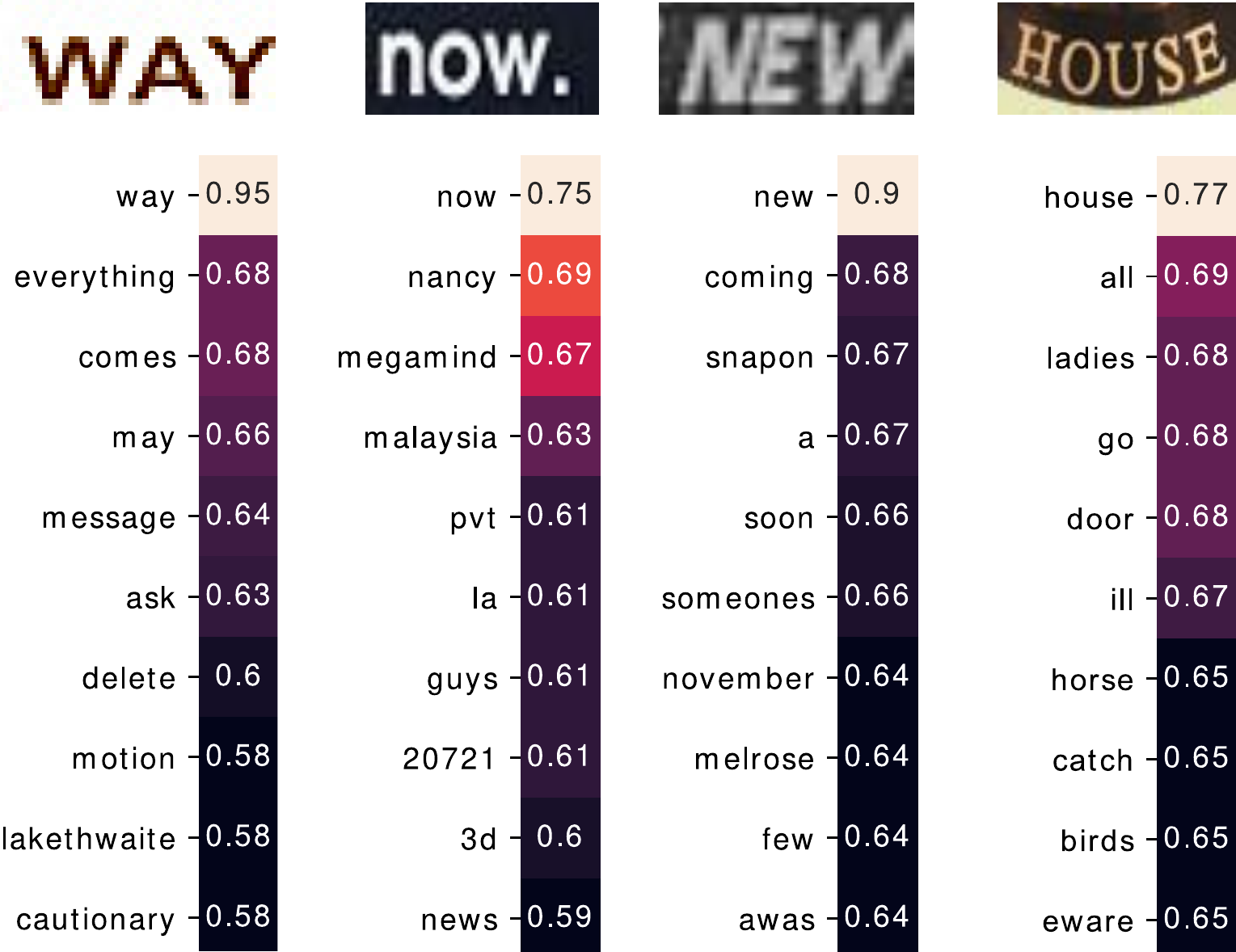}
   }
   \caption{Visualization of cosine similarity of the predicted semantic information from the image w.r.t the word embedding of the words from lexicons. Larger value means more similar semantics. }
   \label{fig_word_embed}
\end{figure}

\begin{table}[h]
\begin{center}
   \begin{tabular}{l|c|c|c|c}
   \hline
   Methods & IC13 & IC15 & SVT & SVTP \\
   \hline
   SAR~\cite{li2019show} & \textbf{91.0} & 69.2 & 84.5 & 76.4 \\
   SE-SAR & 90.9 & \textbf{73.4} & \textbf{85.8} & \textbf{78.7} \\
   \hline
   \end{tabular}
\end{center}
   \caption{Recognition performance on SAR and SE-SAR. }
   \label{tabel_se-sar}
\end{table} 

We conduct some experiments on IC13, IC15, SVT, and SVTP to show the effectiveness of the SE-SAR. The results are demonstrated in Tab.~\ref{tabel_se-sar}. Compared with the baseline, our SE-SAR improves $ 4.2\% $, $ 1.3\% $ and $ 2.3\% $ on IC15, SVT, and SVTP respectively. SE-SAR is only comparable with SAR in that low-quality images are scarce in IC13.

\begin{table*}[t]
\small
\begin{center}
   \begin{tabular}{|l|c|c|c|c|c|c|}
   \hline 
   Methods
   & IIIT5K & SVT & IC13 & IC15 & SVTP & CUTE \\ 
   \hline
   \hline
   Shi \textit{et al.}~\cite{Shi2016An} & 81.2 & 82.7 & 89.6 & - & - & - \\
   Shi \textit{et al.}~\cite{shi2016robust} & 81.9 & 81.9 & 88.6 & - & 71.8 & 59.2 \\
   Lee \textit{et al.}~\cite{lee2016recursive} & 78.4 & 80.7 & 90.0 & - & - & - \\
   Yang \textit{et al.}~\cite{yang2017learning}* & - & - & - & - & 75.8 & 69.3 \\
   Cheng \textit{et al.}~\cite{cheng2017focusing}* & 87.4 & 85.9 & 93.3 & 70.6 & - & - \\
   Cheng \textit{et al.}~\cite{cheng2018aon} & 87.0 & 82.8 & - & 68.2 & 73.0 & 76.8 \\
   Liu \textit{et al.}~\cite{liu2018char}* & 92.0 & 85.5 & 91.1 & 74.2 & 78.9 & - \\
   Bai \textit{et al.}~\cite{bai2018edit}* & 88.3 & 87.5 & \textbf{94.4} & 73.9 & - & - \\
   Liu \textit{et al.}~\cite{liu2018squeezedtext}* & 87.0 & - & 92.9 & - & - & - \\
   Liu \textit{et al.}~\cite{liu2018synthetically} & 89.4 & 87.1 & \underline{94.0} & - & 73.9 & 62.5 \\
   Liao \textit{et al.}~\cite{liao2019scene}* & 91.9 & 86.4 & 91.5 & - & - & 79.9 \\
   Zhan \textit{et al.}~\cite{zhan2019esir} & 93.3 & \textbf{90.2} & 91.3 & 76.9 & 79.6 & 83.3 \\
   Xie \textit{et al.}~\cite{xie2019aggregation} & - & - & - & 68.9 & 70.1 & 82.6 \\
   Li \textit{et al.}~\cite{li2019show} & 91.5 & 84.5 & 91.0 & 69.2 & 76.4 & 83.3 \\
   Luo \textit{et al.}~\cite{luo2019moran} & 91.2 & 88.3 & 92.4 & 74.7 & 76.1 & 77.4 \\
   Yang \textit{et al.}~\cite{yang2019symmetry}* & \textbf{94.4} & 88.9 & 93.9 & \underline{78.7} & \underline{80.8} & \textbf{87.5} \\ 
   \hline
   ASTER~\cite{shi2018aster} & 93.4 & 89.5 & 91.8 & 76.1 & 78.5 & 79.5 \\
   ASTER baseline reproduced & 93.5 & 87.2 & 90.9 & 78.4 & 79.1 & 82.3 \\ 
   SE-ASTER (Ours) & \underline{93.8} & \underline{89.6} & 92.8 & \textbf{80.0} & \textbf{81.4} & \underline{83.6} \\
   \hline
   \end{tabular}
\end{center}
\caption{Lexicon-free performance on public benchmarks. \textbf{Bold} represents the best performance. \underline{Underline} represents the second best result. * indicates using both word-level and character-level annotations to train model.}
\label{tabel_sota}
\end{table*}

\subsection{Qualitative Results and Visualization}

We visualize low-quality images including blur or occlusion. Some examples are shown in Fig.~\ref{fig_low_quality}. As can be seen, our proposed methods SE-ASTER and SE-SAR are robust with low-quality images. We explain that semantic information will provide an effective global feature to decoder, which is robust to the interference in the images. 

We also perform experiments on IIIT5K to visualize the validity of the predicted semantic information. As illustrated in Fig.~\ref{fig_word_embed}, we compute the cosine similarity between the predicted semantic information and the word embedding of each word from lexicons (50 words for each image). In Fig.~\ref{fig_word_embed} (a), the predicted semantic information is very related to the words which have similar semantics. For example, ``home'', ``house'', and ``lodge' all have the meaning of residence. ``Tom'', ``Paul'' and ``Charles'' are all common names. The second row illustrates the robustness of the predicted semantic information. For example, ``house'' and ``horse'' have a similar spelling and are of the edit distance of 1, but their semantics are quite different as shown in Fig.~\ref{fig_word_embed} (b). With the help of global semantic information, the model can distinguish them easily.


\subsection{Comparison with State-of-the-art}

We also compare our methods with previous state-of-the-art methods on several benchmarks. The results are shown in Tab.~\ref{tabel_sota}. Compared with other methods, we achieve 2 best results and 3 second best results out of 6 in the lexicon-free scenario with only word-level annotations. 

Our proposed method works effectively on some low-quality datasets such as IC15 and SVTP compared with other methods. Especially, SE-ASTER improve $ 3.9\% $ on IC15 (from $ 76.1\% $ to $ 80.0\% $) and $ 2.9\% $ on SVTP (from $ 78.5\% $ to $ 81.4\% $) compared with ASTER~\cite{shi2018aster}. It also outperforms state-of-the-art method ScRN~\cite{yang2019symmetry} $ 0.6\% $ on SVTP and $ 1.3\% $ on IC15, although our method is based on a weaker backbone and without character-level annotations.

SE-ASTER also gets superior or comparable results on several high-quality datasets. Compared with ASTER~\cite{shi2018aster} we get $ 0.4\% $ and $ 4.1\% $ improvements on IIIT5K and CUTE respectively. On SVT and IC13, our method gets accuracies of $ 89.6\% $ and $ 92.8\% $, which are slightly worse than ESIR~\cite{zhan2019esir} and~\cite{bai2018edit} by $ 0.6\% $ and $ 1.6\% $. Note that our framework is very flexible and can be integrated with most existing methods, and we believe that if we replace a stronger baseline model better results can be achieved. 


\section{Conclusion and Future Works}
\label{section_5}

In this work, we propose the semantics enhanced encoder-decoder framework for scene text recognition. Our framework predicts an additional global semantic information supervised by the word embedding from a pre-trained language model. Using the predicted semantic information as the decoder initialization, the recognition accuracy can be improved especially for low-quality images. By integrating the state-of-the-art method ASTER into our framework, we can achieve superior results on several standard benchmark datasets. In the future, we will extend our framework to an end-to-end text spotting system. In this way, more semantic information can be utilized.



\section*{Acknowledgment}

This work is supported by the National Key R\&D Program of China (2017YFB1002400) and the Strategic Priority Research Program of Chinese Academy of Sciences (XDC02000000). In addition, we sincerely thank Mingkun Yang for his help.


{\small
\bibliographystyle{ieee_fullname}
\bibliography{egbib}
}

\end{document}